\documentclass[runningheads]{llncs}
\usepackage{graphicx}
\usepackage[numbers]{natbib}
\usepackage{amsmath}
\usepackage{adjustbox}
\usepackage{float}
\usepackage{caption}
\usepackage[misc,geometry]{ifsym}
\usepackage{subcaption}

\usepackage{hyperref}
\hypersetup{
    colorlinks=true,
    linkcolor=blue,
    filecolor=magenta,      
    urlcolor=cyan,
    pdftitle={Overleaf Example},
    pdfpagemode=FullScreen,
    citecolor=blue,
    allcolors=blue
    }

\begin{document}
\title{Application of Federated Learning Techniques for Arrhythmia Classification Using 12-Lead ECG
Signals}
\author{Daniel Mauricio Jimenez Gutierrez (\Letter) \orcidID{0000-0002-0305-814X} \and
Hafiz Muuhammad Hassan\ \and
Lorella Landi \and
Andrea Vitaletti\orcidID{0000-0003-1074-5068}  \and
Ioannis Chatzigiannakis\orcidID{0000-0001-8955-9270}
}

\authorrunning{D.~Jimenez, H.~Hassan, L.~Landi, A.~Vitaletti, I.~Chatzigiannakis}% Part of LEFT running header
\titlerunning{Application of FL techniques for 12-lead ECG arrhythmia classification}% Part of RIGHT running header
% \author[1]{Daniel Mauricio Jimenez Gutierrez}

% \author[1]{Hafiz Muuhammad Hassan}

% \author[1]{Lorella Landi}

% \author[1]{Andrea Vitaletti}[orcid= 0000-0003-1074-5068]

% \author[1]{Ioannis Chatzigiannakis}[orcid= 0000-0001-8955-9270]
% \ead{ichatz@diag.uniroma1.it}

\institute{Sapienza University, Rome, Italy \\
\email{jimenezgutierrez@diag.uniroma1.it (\Letter), vitaletti@diag.uniroma1.it, ichatz@diag.uniroma1.it}}
% , hassan.1873829@studenti.uniroma1.it, landi.1912333@studenti.uniroma1.it

%
\maketitle              % typeset the header of the contribution
\begin{abstract}
Artificial Intelligence-based (AI) analysis of large, curated medical datasets is promising for providing early detection, faster diagnosis, and more effective treatment using low-power Electrocardiography (ECG) monitoring devices information. However, accessing sensitive medical data from diverse sources is highly restricted since improper use, unsafe storage, or data leakage could violate a person's privacy. This work uses a Federated Learning (FL) privacy-preserving methodology to train AI models over heterogeneous sets of high-definition ECG from 12-lead sensor arrays collected from six heterogeneous sources. We evaluated the capacity of the resulting models to achieve equivalent performance compared to state-of-the-art models trained in a Centralized Learning (CL) fashion. Moreover, we assessed the performance of our solution over Independent and Identical distributed (IID) and Non-IID federated data. Our methodology involves machine learning techniques based on Deep Neural Networks and Long-Short-Term Memory models. It has a robust data preprocessing pipeline with feature engineering, selection, and data balancing techniques. Our AI models demonstrated comparable performance to models trained using CL, IID, and Non-IID approaches. They showcased advantages in reduced complexity and faster training time, making them well-suited for cloud-edge architectures.

\end{abstract}
%
%
%
%%%%%%%%%%%%%%%%%%%%%%%%%%%%%%%%%%%%%%%%%
%%%%%%%%%%%%%%%%%%%%%%%%%%%%%%%%%%%%%%%%%
\section{Introduction}
\label{sec:intro}
Significant advances in Artificial Intelligence (AI) based medical data analysis, i.e., big data, machine learning (ML), deep learning (DL), etc., drive innovation in MedTech~\cite{hamet2017artificial,muller2022explainability}. Numerous studies demonstrate how scalable data digitization, combined with data technologies, can significantly contribute to timeliness and resource effectiveness of prevention, diagnosis, and treatment~\cite{arnold2017doctor}. Modern AI models feature millions of parameters that need to be learned from sufficiently large, curated datasets to achieve clinical-grade accuracy while being safe, fair, equitable, and generalizing well to unseen data~\cite{wang2019deep}. Recent publications provide evidence on how large volumes of data managed by independent and possibly heterogeneous \textit{low-power devices}, often on a cross-border basis, are shared to drive AI systems that can outperform specialists or, at the very least, help them reduce their effort significantly~\cite{mckinney2020international}. Despite its advantages, the practical application of this technological evolution in healthcare environments faces certain difficulties~\cite{panch2019inconvenient}: the rate at which AI technologies are being developed is out of step with the rate at which they are being adopted in practice. The integration of AI in healthcare operations and services creates a solid and urgent demand to integrate medical data obtained from various edge devices~\cite{he2019practical}. However, since health data is susceptible and its use is strictly restricted, data like this is challenging to obtain~\cite{van2014systematic}. Indeed, this is exacerbated in the case of personal information, where improper use, unsafe storage, data leakage, or abuse could violate a person's privacy. Even if data anonymization could circumvent these limits, it is now widely accepted that deleting metadata such as a patient's name or date of birth is frequently insufficient to maintain privacy~\cite{rocher2019estimating}.

Recently, Federated learning (FL) has been introduced as a decentralized training paradigm of an AI model that enables collaborative learning without exchanging the data itself~\cite{yang2019federated}. FL utilizes distributed consensus algorithms that allow organizations to gain insights collaboratively so that patient data are not transferred outside the organization's firewalls. Consider, for example, multiple hospitals (equipped with low-powered medical devices) collaboratively learning from a shared medical predictive model while keeping data sources in their original location in each hospital, decoupling the ability to do ML from the need to store the data in a single place. Some recent studies examine the performance levels achieved by AI models trained using FL techniques on isolated single-organizational data. They compare them with models trained on centrally hosted data, indicating comparable results~\cite{sheller2018multi}. Therefore FL offers a promising path with a potentially significant impact on large-scale precision medicine, resulting in models that make impartial judgments, best reflect an individual's physiology, and are sensitive to uncommon illnesses while respecting governance and privacy issues.

% In this work, we employ FL to efficiently train AI models for arrhythmias monitoring, a common cardiovascular disease. Arrhythmia refers to irregularities in heart rate or rhythm and is a leading global cause of death. Low-power Electrocardiography (ECG) monitoring devices utilize multiple sensors called ECG leads, primary tools for identifying heart rhythm variations~\cite{murat2020application}. 

In this work, we apply the FL paradigm to train AI models that can assist doctors in closely and efficiently monitoring arrhythmias, a common Cardiovascular disease (CVDs) recognized as one of the leading causes of death globally\footnote{\url{https://www.who.int/en/news-room/fact-sheets/detail/cardiovascular-diseases-(cvds)}}. Arrhythmia is a disorder of the heart rate (pulse) or heart rhythm, being too slow, fast or irregular. Today, the most effective tool to identify variations in heart rhythm along with other patterns of the heart's electrical impulses is based on low-power Electrocardiography (ECG) monitoring devices~\cite{murat2020application}. Such devices use multiple sensors, and the ECG leads to capture the heart's electrical activity from different angles and positions. Automatic analysis of the ECG signal has become one of the essential tools in diagnosing heart disease. It has been used as the basis for the design of algorithms for many years~\cite{LUZ2016144}. 

We introduce a multi-step privacy-preserving methodology for training AI models for arrhythmia detection and classification using 12-lead ECG recordings. The process can be applied to various ML techniques, ranging from deep neural networks to recurrent neural networks with long short-term memory. Moreover, the methodology is generic enough to be applied to several classification problems. A detailed feature extraction stage is included that examines the morphology of all the channels of the ECG signals both in the time domain and the frequency domain and the cardiac rhythm of consecutive heartbeats, resulting in about 650 features. The methodology introduces a novel feature selection stage where a subset of the features is selected so that the overall training time of the model and performance achieved is optimized.

We apply the FL paradigm over the  PhysioNet in Cardiology Challenge 2020 dataset~\cite{physionet_challenge_2020}, which includes high-definition 12-lead ECG recordings obtained from $43,059$ patients recruited by six disparate geographical centers. In contrast to other publicly available datasets of ECG recordings, the one considered here (a) provides the most extensive set of cardiac abnormalities, thus contributes significantly to training AI models that are capable of recognizing a broad range of cardiac arrhythmias; (b) the recordings provided by the six centers are heterogeneous as they include a different number of patients and with different distribution of arrhythmias. We evaluate the capacity of the resulting AI models to achieve equivalent performance when compared to state-of-the-art models trained (a) when the same data is collected in a \textit{central place} -- this allows us to study the performance following the currently dominant, centralized training approach; (b) when the data is available at \textit{multiple homogeneous devices}, that it is Independent and Identically Distributed among the devices (IID) -- this allows us to study the performance in a simplified federated environment; (c) when the data is available at \textit{multiple heterogeneous devices}, that it is non-independent and identically distributed among the devices (Non-IID) -- this is the most realistic scenario since the distribution of the different types of arrhythmias recorded varies between hospitals.

%%%%%%%%%%%%%%%%%%%%%%%%%%%%%%%%%%%%%%%%%
%%%%%%%%%%%%%%%%%%%%%%%%%%%%%%%%%%%%%%%%%
\section{Previous related work}
\label{sec:sota}
\subsection{Arrythmia detection}

During the past years, the development of AI models has been based on free access ECG datasets, such as PhysioNet~\cite{932728}. Probably the most diffused dataset for ECG signal analysis and arrhythmia detection among all those freely available is the  \textit{Massachusetts Institute of Technology-Beth Israel Hospital (MIT-BIH) Arrhythmia database}~\cite{moody2001impact}. The latter is because, first, the dataset includes recordings of all five arrhythmia classes suggested by the AAMI standards\cite{aami}. Second, as two cardiologists annotate all the recordings included, it makes it suitable for supervised and unsupervised learning techniques. On the other hand, the limitations of the MIT-BIH dataset in terms of (a) signal quality, since only two ECG leads were used during the recordings, and (b) dataset size, since the different types of arrhythmias identified on the $48$ patients, inevitably restrict the generality of the trained AI-models.

In recent years, machine-learning methods have rapidly detected cardiac abnormalities in 12-lead ECGs \cite{arythmia_detection_1, arythmia_detection_2}. Newly emerging deep-learning models have further achieved comparable performance to clinical cardiologists on many ECG analysis tasks \cite{arythmia_detection_3} and arrhythmia/disease detection  \cite{arythmia_detection_6}. However, as high-quality, real-world ECG data is challenging to acquire, most deep-learning models are designed to detect only a tiny fraction of cardiac arrhythmias, owing to the limitations of the datasets. In addition, due to the database's structure and annotations, many of those \cite{sahoo2017multiresolution, lin2014heartbeat} focus only on heartbeat classification. Most of the ones \cite{rajkumar2019arrhythmia, rohmantri2020arrhythmia} that use the MIT-BIH database for rhythmic arrhythmia classification, on the other side, only deal with a small number of arrhythmias, so their good performance results are not comparable with those of this paper due to the limited range of arrhythmias detected and the differences in database dimensions.

To encourage more multidisciplinary research, PhysioNet in Cardiology Challenge 2020 (Challenge 2020) \cite{physionet_challenge_2020} provided high-quality 12-lead ECG data obtained from multiple centers with a large set of cardiac abnormalities. The aim of Challenge 2020 was to identify clinical diagnoses from 12-lead ECG recordings, providing an opportunity to employ various advanced methods to address clinically important questions that are either unsolved or not well-solved \cite{arythmia_detection_9}.

Forty-one teams got selected during this challenge, and others were discarded due to failing to achieve high scores or appear at the conference. One can observe that most teams use common techniques by looking at the 41 teams' papers published in PhysioNet \cite{physionet_challenge_2020}. Among these techniques, signal processing, DNNs, convolutional neural networks (CNNs), and end-to-end and multi-binary classifications are used by all of the top 10 teams. In addition, there are several important points 1) deep-learning methods were more popular than traditional methods in Challenge 2020; 2) all the teams that employed deep-learning methods used CNNs; and 3) none of the top 10 teams used hand-labeled features (except demographic features); they all adopted end-to-end models instead. 

One can also notice that the three highest-ranking teams used the model ensemble \cite{first_team, second_team, third_team}, but only 14 out of 41 teams employed this strategy. It is also important to note that model ensemble only helps if used for a single model rather than structurally different models. Most of the team also used only age and sex as features rather other using demographic features or 12-lead ECG-based features. The training data in Challenge 2020 suffer from heavy class imbalance, so most teams used threshold optimization \cite{eighth_team, second_team, ninth_team} and weighted loss \cite{sixth_team, seventh_team} to handle the imbalance class issue. In addition, over-sampling \cite{thirteen_team}, down-sampling \cite{sixteen_team}, and other methods have been employed in Challenge 2020.

\subsection{Federated learning}

FL applied to health has gained interest in the last few years. Specifically, employing FL to classify ECG arrhythmias shows relevant advances and results \cite{fl_previouswork3}. For example, \cite{fl_previouswork} implemented explainable artificial intelligence (XAI) and convolutional neural networks (CNN) for ECG-based healthcare. They employed the baseline Massachusetts Institute of Technology - Boston’s Beth Israel Hospital (MIT-BIH) Arrhythmia database. The trained classifier obtained an accuracy of up to 94.5\% and 98.9\%.

Utilizing the same MIT-BIH, \cite{fl24} proposed an algorithm to identify the best number of epochs per training round in FL. They showed that using its algorithm's suggested number of epochs decreases the training time and resource consumption while keeping the accuracy level. In detail, their approach obtained an accuracy close to 97.5\%.

Using ECG collected from different medical institutions, the authors of \cite{fl_previouswork2} obtained a dataset where the arrhythmia's distribution is diverse from one hospital to another. The latter may lead to the non-convergence of the FL-based algorithm. To address that challenge, they optimized the FL-based algorithm using a sharing strategy for partial ECG data of each medical institution combined with the elastic weight consolidation (EWC) algorithm. As a result, their model's performance measured through the F1-Score got 0.7.

FL can be classified into different types depending on how the data is divided among the devices (data partitioning) \cite{yang2019federated}. One of those classifications is the so-called \textbf{Horizontal data partitioning}, in which the data of various local nodes share the same properties (i.e., features). However, there is a limited sample space intersection (i.e., the patients differ from device to device). Moreover, the most common FL architecture is horizontal partitioning \cite{horizontal_fl}.

FL has relevant advantages over centralized (traditional) systems \cite{fl27_adv_disadv1}. The most remarkable is that the data remains locally (in each hospital) and does not require transmitting personal information along the network. It only sends the weights of the models. A second advantage is that models continuously enhance by utilizing local nodes' input, eliminating the requirement to aggregate data for continual learning. In third place, FL generates technological efficiency. This technique employs less complex hardware since FL models do not require a single central server to train data.

FL, on the other hand, faces several difficulties. Firstly, models could be poisoned by providing model updates resulting from mislabeled data, even though only models, not raw data, are sent to the central server. FL is vulnerable to adversarial assaults in the form of backdoors during training, as seen in~\cite{fl27_adv_disadv2}. A backdoor aims to tamper with the trained model's performance on specified sub-tasks. Secondly, device-specific characteristics may prevent models from generalizing from some local nodes, decreasing the global model's performance \cite{fl27_adv_disadv1}. The mentioned difficulties and how to tackle them are out of the scope of this work.

%%%%%%%%%%%%%%%%%%%%%%%%%%%%%%%%%%%%%%%%%
%%%%%%%%%%%%%%%%%%%%%%%%%%%%%%%%%%%%%%%%%
\section{Methodology}
\label{sec:methods}
In this study, we assume \textit{a group of healthcare organizations (equipped with low-powered medical devices)} that wish to collectively train an arrhythmia classification AI-based module without sharing their medical records. For each organization, we envision a \textit{single, local module} that coordinates all the activities related to collecting, storing, and analyzing medical records. Moreover, we assume that each organization can access high-definition electrocardiogram monitoring devices to record the patient's heart activity. Each monitoring session produces a short 12-channel ECG recording, i.e., about 16 seconds, that is transferred from the device to the local, private database of the organization\footnote{Remark that several different interconnection architectures are used in ECG technologies available in the market and studied in the relevant literature, such as wireless technologies like WI-FI or BLE, or wire technologies such as USB, or non-volatile memory formats. Such interconnections aspects are beyond the scope of this paper.}. One or more healthcare experts examine the ECG recordings and provide a diagnosis stored in the local database.

We assume that the healthcare organizations have agreed upon a \textit{single, global, trusted cloud server}. The role of the global server is to coordinate all the activities of the local modules. The global server does not maintain any database with medical records. The only information stored is related to the standard AI model and the system's operating parameters. We also assume that all organizations have agreed on a standard length for the ECG recordings, e.g., $16$ seconds, and a common sampling frequency, e.g., $257KHz$\footnote{Note that in case the recording has a different sampling frequency, various algorithms exist in the relevant literature to change the sampling frequency to a lower or a higher one without affecting the accuracy. \cite{competence_group2}}$^{,}$\footnote{Note that ECG recordings that are longer than the agreed length can be split into multiple ones without loss of generality.}.

\begin{figure*}[ht]
\centering
\includegraphics[width=\textwidth]{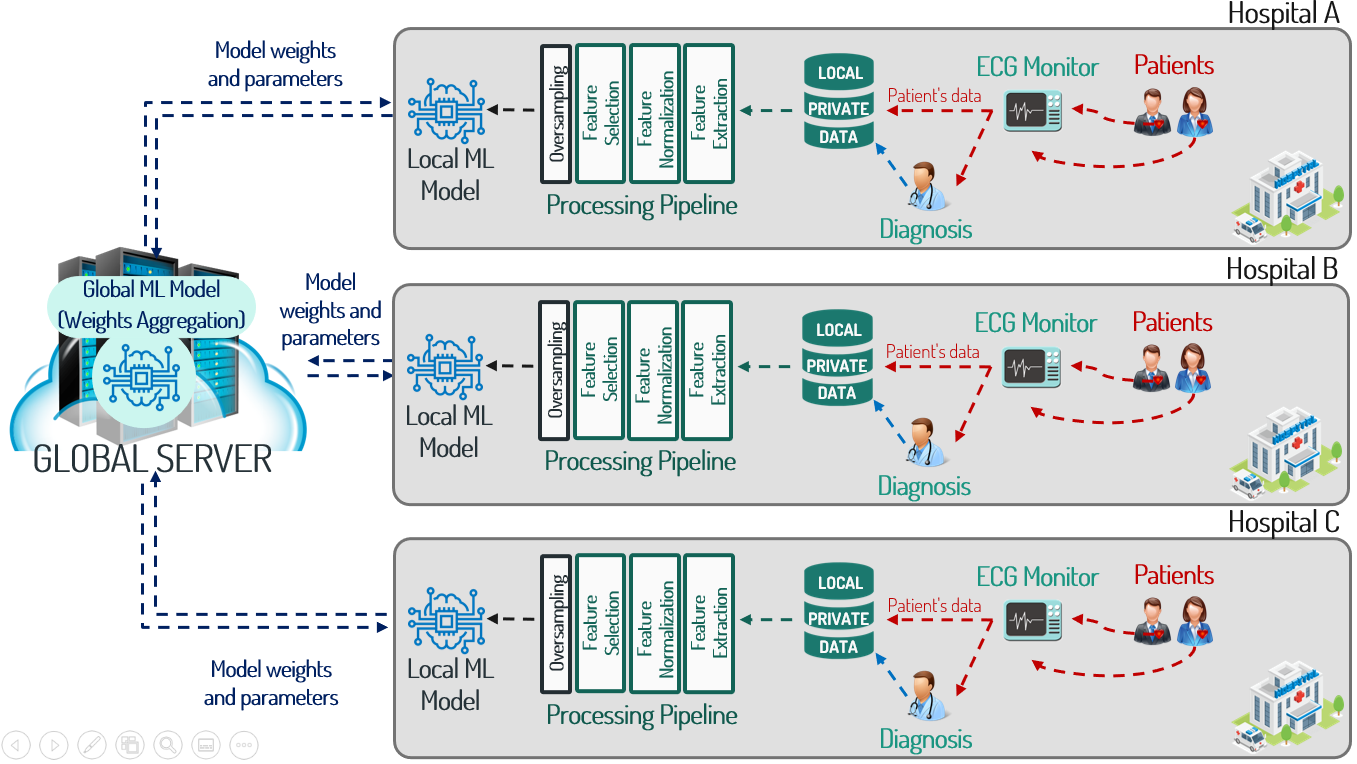}
\caption{High-level overview of the proposed federated learning methodology and software and hardware components.}
\label{fig:fl_architecture_global}
\end{figure*}

Periodically, the global server starts \textit{a global training session} by notifying all the local servers (a.k.a, clients or local nodes) of the organizations that participate in the federation. Upon receiving this notification, each organization's device independently goes through \textit{a local training session}. During a local training session, all the records available in the local, private database are analyzed using local computing resources based on a \textit{a processing pipeline} made up of four steps. First, each recording passes through a \textit{Feature Extraction} phase, where they are analyzed independently, and critical information connected to the heartbeats that will help the training of the AI model is extracted. Second, a \textit{feature normalization} step follows, where statistics are used to scale the features to improve the robustness of the data. In coordination with the global server, the local servers compute the necessary robust measures over the federated dataset without revealing sensitive information. Third, all the original records, along with the normalized features, are examined. To this end, a \textit{feature selection} is carried out to remove redundant features that may hinder the performance of the trained model and reduce the computational cost since they use a smaller number of features to train the local model. The fourth and final \textit{data balancing} step examines the diagnosis attached to all the records in the local database to identify and remove any imbalances found between the representation of arrhythmia classes, thus increasing the generalization power of the model.

When all the local servers have completed the processing pipelines, under the coordination of the global server, they start training their local models. When the training is concluded, the weights of the resulting model are transmitted to the global server. The global server examines all the individual weights using a \textit{weight aggregation method} and sends the resulting model to all organizations. This process is repeated until either the \textit{distributed optimization converges} or a certain number of steps is reached.
 
A high-level overview of the above methodology and the interconnection of the software and hardware elements that make up the federated architecture are depicted in Fig.~\ref{fig:fl_architecture_global}.

Regarding the data partitioning, the FL type shown in Fig.~\ref{fig:fl_architecture_global} is a Horizontal FL since each local node collects the same features. Additionally, regarding the ML models, the FL approach exposed is homogeneous because each hospital's device applies the same model with different data. Finally, regarding scale, the FL defined is known as cross-device because the number of local nodes is high, but they do not have enough computing power~\cite{fl25}.

\subsection{Problem Formulation}

We assume that an ECG recording $s_i$ has $c \in [1, 12]$ channels with fixed samplings rate, producing $d$ samples per channel, $s_i = \left\{x_1^1 \ldots x_d^1, x_1^2 \ldots x_d^2, \ldots,  x_1^{12} \ldots x_d^{12} \right\}$ where $x_j^c$, is the $j^{th}$ sample of channel $c$, where $j \in [1, d]$. The ECG multiclass arrhythmia classification task considered here takes as input a local data set of $n$ 12-channel ECG recordings (${\cal S} = \{s_1 , \ldots, s_n \}$). It outputs a sequence of labels, one for each signal ${\cal L} = [l_1 , \ldots, l_n ]$, where each $l_i \in {\cal A}$, and ${\cal A}$ represents the different rhythm classes considered.  We focus on $|{\cal A}| = 27$ diagnoses of arrhythmia classes as defined by the AAMI standard~\cite{aami} that are of clinical interest and more likely to be recognizable from ECG recordings~\cite{phy2020}. Table~\ref{table:selectedCategories} summarizes the complete list of diagnoses considered.

\begin{table}[ht]
\centering
\resizebox{7cm}{!}{
\begin{tabular}{ |p{5.2cm}|p{2.2cm}|}
\hline
\textbf{Diagnosis} & \textbf{Abbreviation}\\ 
 \hline\hline
1st degree AV block	& IAVB \\\hline
Abnormal QRS & abQRS \\\hline
Atrial fibrillation	& AF \\\hline
Left atrial enlargement	& LAE \\\hline
Left axis deviation	& LAD \\\hline
Left bundle branch block & LBBB \\\hline
Left ventricular hypertrophy & LVH \\\hline
Low QRS voltages & LQRSV \\\hline
Myocardial infarction & MI \\\hline
Myocardial ischemia & MIs \\\hline
Nonspecific st t abnormality & NSSTTA \\\hline
Old myocardial infarction & OldMI \\\hline
Pacing rhythm & PR \\\hline
Premature atrial contraction & PAC \\\hline
Prolonged QT interval & LQT \\\hline
Q wave abnormal & QAb \\\hline
Right bundle branch block & RBBB \\\hline
Sinus arrhythmia & SA \\\hline
Sinus bradycardia & SB \\\hline
Sinus rhythm & NSR \\\hline
Sinus tachycardia & STach \\\hline
ST depression & STD \\\hline
ST elevation & STE \\\hline
ST interval abnormal & STIAb \\\hline
T wave abnormal & TAb \\ \hline
T wave inversion & TInv \\ \hline
Ventricular ectopics & VEB \\ \hline
\end{tabular}}
\caption{Diagnoses and abbreviations for the selected diagnoses}
\label{table:selectedCategories}
\end{table}

\subsection{Feature Extraction}

\begin{figure}[ht]
\centering
\includegraphics[width=0.6\columnwidth]{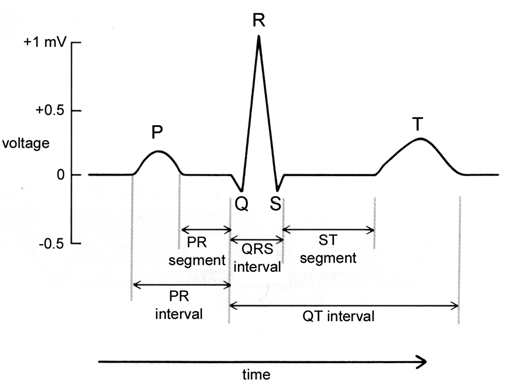}
\caption{Typical ECG waveform from the Lead II position for one cardiac cycle.}
\label{fig:ECG_waveform}
\end{figure}

First, each 12-channel ECG recording is analyzed by examining the morphology of all the heart cycles included. The QRS complex represents a heart cycle, illustrated in Fig.~\ref{fig:ECG_waveform}, which corresponds to the depolarization of the right and left ventricles of the human heart. Detecting the QRS complex is essential for time-domain signal analysis, that is, the heart rate variability. For each found QRS complex, a series of representative \textit{morphological features}\footnote{Code for morphological features: \url{https://github.com/physionetchallenges/python-classifier-2020/blob/master/get_12ECG_features.py}} is extracted in terms of slope, amplitude, and width, such as the \textit{R peak}, the highest amplitude of the complex;\textit{PR interval}, the period between the P peak and the R peak; \textit{QRS interval}, the period between the Q and the S segments; \textit{ST interval}, the period between the S and T segments; \textit{QT interval}, the period between the Q and T segments;  the \textit{RR interval}, the period between two consecutive R peaks. 

An alternative type of feature based on \textit{Spectral analysis} was employed in this work. The latter is a frequency-based signal analysis that uses a Discrete Wavelet Transform (DWT) followed by spectral analysis\footnote{Code for spectral features: \url{https://github.com/onlyzdd/ecg-diagnosis/blob/dfa9033d5ae7be135db63ff567e66fdb2b86d76d/expert_features.py}}. For each of the coefficients of the DWT, as well as the original complete signal, the percentiles (5, 25, 50, 75, 95), mean, standard deviation, variance, skewness, and kurtosis, are calculated along with Shannon’s Entropy (information of the distribution). The DWT coefficients were obtained using the \textit{wavedec} function of the \textit{pywt} Python package. The age and sex of patients are also included as features, resulting in a total of 650 features (14 morphological and demographical and 636 spectral). Note that this step is done only once for each record, and the extracted features are stored on the local database for future training sessions.

\subsection{Feature Normalization}
\label{sec:norm}

Towards enhancing the performance of the AI models, the values of each feature are centered and scaled independently according to the median value and the interquartile range (IQR) between the first and third quartiles, i.e., the 25th and 75th percentiles, using the \textit{RobustScaler} Python function. The specific formula is as follows:
\begin{equation} \label{robustscaler_eq}
\mbox{norm}\left({}_{f}x_{i}^{c}\right) = \frac{{}_{f}x_i^c - \bar{{}_{f}x^c}}{{}_{f}x_{q_{75}}^c - {}_{f}x_{q_{25}}^c}
\end{equation}
Where $f$ is one of the 636 spectral features, $\bar{{}_{f}x^c}$ is the median, ${}_{f}x_{q_{35}}^c$ and ${}_{f}x_{q_{75}}^c$ are the 25th and 75th percentiles computed over all the values in the entire federated data set for the given feature $f$ in channel $c$.

\subsection{Feature Selection}

Since the extraction of the majority of the features is conducted over each of the 12 channels and due to the nature of multi-channel ECG recording, in many cases, there is redundancy between pairs of features. Moreover, environmental noise from nearby appliances or noise originating from muscular activity frequently produces abnormalities in a recording identified and removed by the ECG monitor at the hardware level~\cite{serhani2020ecg}. The latter leads to temporal loss of some values of one or more of the 12 channels~\cite{mariappan2016effects}. Since redundant or missing features can potentially lower the performance of the classification model, it is crucial to identify and remove them. Besides improving our model's predictive performance, it will also reduce the overall computational requirements of AI model training~\cite{dhal2022comprehensive}. 

Several methods have been proposed for feature selection in the relevant bibliography~\cite{guyon2003introduction}. Here we offer a simple and automated way to rank features based on their importance. We calculate the embedded feature importance scores based on tree ensemble models. The central concept here is that given the feature sub-sampling and bootstrapping in an ensemble of trees, the importance scores of two or more redundant features are expected to be spread evenly among the trees. Then the resulting scores of the candidate features are re-examined following a sequential boosting process based on the XG-Boost model~\cite{10.1145/2939672.2939785}. The boosting process allows re-examining some of the features that received a low rank due to specific limitations of the tree-based feature importance calculation in large datasets, datasets with missing values, and datasets with both numerical and categorical features~\cite{ALSAHAF2022115895}.

\subsection{Dataset Balancing}
Various nationwide register-based cohort studies conducted over a very long period, such as in Denmark~\cite{wodschow2021geographical} and Canada~\cite{rosychuk2015geographic}, indicate an inequality of arrhythmia incidence rate between highest and lowest income groups and the existence of significant geographical variation even in countries with free access to healthcare and even when accounting for socioeconomic differences at an individual level. Therefore when the organizations do not share their medical records, the observed distribution of the local diagnoses of the arrhythmia classes will differ across each healthcare organization. As each organization is training its local model based only on the data locally available, it is expected that the resulting model will have certain biases towards the most populated classes, learning too much about their characteristics and failing to recognize the differentiating factors of the others, resulting in poor performance for the minority classes~\cite{he2009learning}.

Various techniques have been proposed to overcome this problem in the relevant literature, such as the Random Over Sampling (ROS)~\cite{ling1998data}, Random Under Sampling (RUS)~\cite{he2009learning}, and Synthetic Minority Oversampling Technique (SMOTE-SMT)~\cite{chawla2002smote}. The approaches implemented were ROS + RUS and SMT + RUS. The latter aimed to overcome the well-known problem of ROS resulting from the generation of exact copies of the minority class samples.

In more detail, for a local dataset made up of $n$ recordings and diagnoses represented as ${\cal D}=\{(s_1, l_1),\dots,(s_n, l_n)\}$. Then $b_u$ is the number of signals belonging to the under-balanced classes, and $b_o$ is the number of over-balanced classes. In this case, $b_u \leq b_o$. Clearly $b_u+b_o=n$. Our balancing method will make a total of $\tau =(m_l - m_s)\cdot \beta$ random duplications (or interpolations) and eliminations, where $\beta \in [0,1]$, and $m_l$ and $m_s$ represent the numbers of recordings in the largest class and the smallest class, respectively. Here $\beta$ is user-defined. For example, if $\beta=1$, the resulting dataset is completely balanced.

The proposed algorithm uses an appropriate mix of random duplication (or interpolation) of the minority, under-balanced class samples, and random elimination of the majority, over-sampled classes. Thus, after identifying the changes that need to be made, the algorithm randomly chooses a sample $x_i$. If it belongs to an under-balanced class, it will duplicate it, while if it belongs to an over-balanced class, it will eliminate it. The resulting approach is simple to implement and has shallow requirements in terms of computational resources.

\subsection{Multiclass Arrhythmia Classification}
\subsubsection{Model 1 - (DNN) Deep Neural Networks}

In this context, deep neural networks (DNN) refer to organic or artificial systems of neurons. DNN can adapt to changing input and produce the best possible result without modifying the output criteria because they can adjust to variable inputs \cite{ann2}. DNN, an artificial intelligence-based concept, has gained popularity for developing health classification systems.

\begin{figure}[ht]
\centering
\includegraphics[width=0.7\columnwidth]{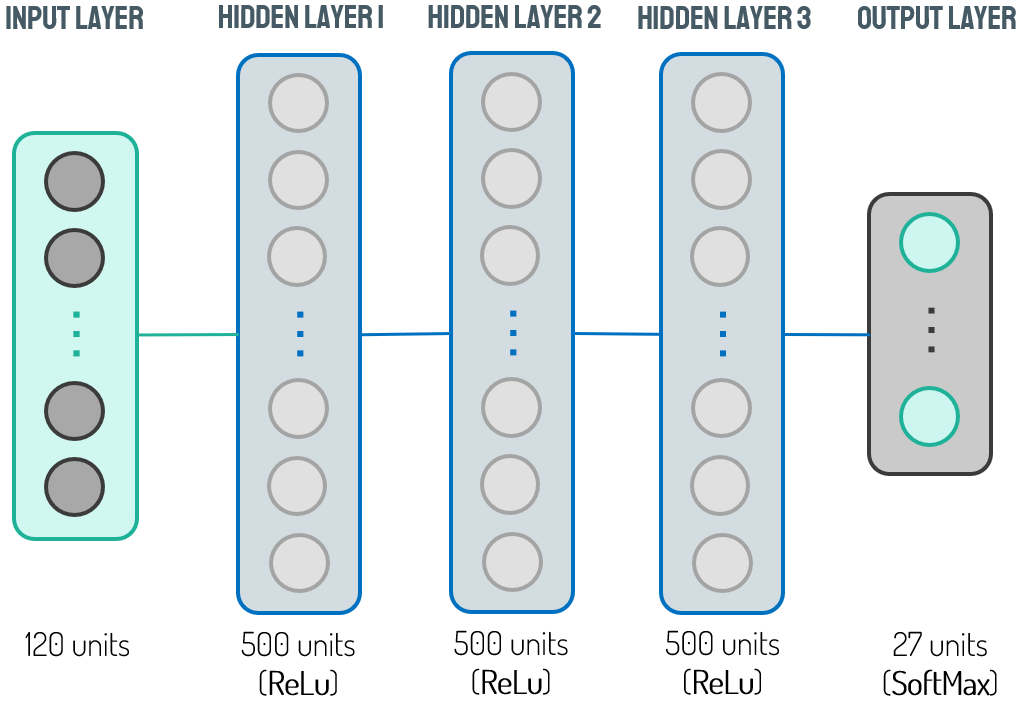}
\caption{DNN architecture}
\label{fig:DNN_architecture}
\end{figure}

As depicted in Fig.~\ref{fig:DNN_architecture}, the DNN defined, inspired by~\cite{inspiration_DNN}, has one input layer, three hidden layers, and one output layer. For the input layer, 120 units were considered since they represent the number of features used in the training set. The three hidden layers contain 500 hidden units each. Furthermore, the last layer was formed by 27 neurons, related to the number of diagnoses to predict. Moreover, the hidden layers employed the ReLu activation and the SoftMax for the output layer. The hidden layers, neurons, and activation functions arose after the hyperparameter fine-tuning method, using the possibilities in Table~\ref{table:gridDNN}. In detail, one model was trained for every combination of parameters. Then, Accuracy, Precision, Recall, and F1-Score were calculated over the test set and the fitted model. Finally, the best approach ended up as the model obtaining the highest F1-Score among all the combinations.

\begin{table}[ht]
\begin{center}
\begin{tabular}{ ||p{5.4cm}||p{3.0cm}||p{2.8cm}|| }
 \hline
\textbf{Hyperparameter} & \textbf{Possible values} & \textbf{Best approach}\\ [0.4ex] 
 \hline\hline
 Number of hidden layers & 2 ,3, 5, 10 & 3 \\
\hline
Number of hidden neurons & 50, 100, 500, 1000 & 500\\
\hline
Activation function (Hidden layers) & ReLu, Tanh, SeLu & ReLu \\
\hline
Activation function (Output layer) & Sigmoid, SoftMax & SoftMax \\
\hline\hline
\end{tabular}
\end{center}
\caption{Fine-tuning grid for DNN}
\label{table:gridDNN}
\end{table}

\subsubsection{Model 2 - (LSTM) Long-Short Term Memory}

Long-short-term memory (LSTM) networks are a type of DNN. It is a Recurrent Neural Networks (RNNs) class that can learn long-term dependencies and functionals to solve sequence prediction issues. Besides, for single data points like pictures, LSTM has feedback connections, which means it can process a complete data sequence~\cite{lstm}. 

\begin{figure}[ht]
\centering
\includegraphics[width=0.5\columnwidth]{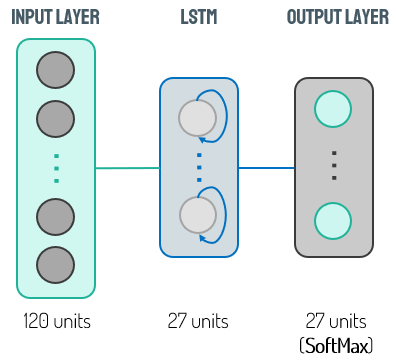}
\caption{LSTM architecture}
\label{fig:lstm_architecture}
\end{figure}

Fig.~\ref{fig:lstm_architecture} depicts the architecture used for the LSTM methodology, motivated by~\cite{inspiration_DNN}. Then, 120 neurons were employed in the input layer because those variables were used to predict. Moreover, both the LSTM cell and the output layer were composed of 27 neurons, as many as the number of classes predicted. Regarding the hyperparameters mentioned, their values were selected after the fine-tuning process, providing the best performance. Table~\ref{table:gridLSTM} depicts the possibilities tested during the hyperparameter fine-tuning procedure. Notice that the number of LSTM cell neurons has an upper limit of 27, corresponding to the number of classes (arrhythmias) considered in this work.

\begin{table}[ht]
\begin{center}
\begin{tabular}{ ||p{5.4cm}||p{2.8cm}||p{2.8cm}|| }
\hline
\textbf{Hyperparameter} & \textbf{Possible values} & \textbf{Best approach}\\ [0.4ex] 
\hline\hline
Number of LSTM cell neurons & 1, 5, 10, 20, 27 & 27\\
\hline
Activation function (LSTM layer) & None, ReLu, Tanh & None \\
\hline
Activation function (Output layer) & Sigmoid, SoftMax & SoftMax \\
\hline\hline
\end{tabular}
\end{center}
\caption{Fine-tuning grid for LSTM}
\label{table:gridLSTM}
\end{table}

\subsection{Distributed Weight Aggregation}

When the weights from each local model arrive at the global node, they are aggregated. Federated Average Aggregation (FedAvg) is typically used for that task \cite{fedavg_fl}. The latter enables local nodes to perform multiple batch updates on local data, exchanging averaged updated weights rather than raw data~\cite{fedavg2}.

In an FL environment exists a concept so-called \textbf{communication round} (comm round)~\cite{comm_rounds2}. The latter begins when a model is trained inside each local node, and later the models' weights are passed to the global node to be aggregated there. Finally, the communication round finishes when each local model gets updated with the new weights. Then, as a result, in each comm round, the model's performance increases. Moreover, it should get stable after some trials~\cite{comm_rounds1}.

The training process in the proposed architecture works as follows. First, each local node trains the model using the correspondent local data. In the second place, every client sends the weights of the obtained model to the global server. The weights are aggregated and sent back to each client in the latter. Finally, each local node updates its parameters. The previous process is repeated until the performance of the global model converges.

%%%%%%%%%%%%%%%%%%%%%%%%%%%%%%%%%%%%%%%%%
%%%%%%%%%%%%%%%%%%%%%%%%%%%%%%%%%%%%%%%%%
\section{Experimental settings and fine-tuning}
\label{sec:exp}

\subsection{Data}
We utilized the 2020 PhysioNet Challenge dataset~\cite{phy2020} to classify cardiac arrhythmias from ECG records, consisting of 12-lead ECG recordings from 43,059 patients. Furthermore, it adds data from different hospitals, making it suitable for an FL approach to address privacy concerns. It offers heterogeneous data and a wide range of 82 cardiac arrhythmias, representing a realistic scenario. The data are from five sources summarized in Table~\ref{table:datasetsSummary}.

\begin{table*}[ht]
\begin{center}
% \resizebox{8cm}{!}{
\begin{tabular}{ ||p{2.9cm}||p{1.7cm}||p{1.7cm}||p{1.4cm}||p{1.4cm}||p{1.7cm}||}
 \hline
\textbf{Dataset} & \textbf{Number of Recordings} & \textbf{Mean Duration (seconds)} & \textbf{Mean Age (years)} & \textbf{Sex (male / female)} & \textbf{Sample Frequency (Hz)}\\ [0.4ex] 
 \hline\hline
CPSC (all data) & 13256 & 16.2 & 61.1 & 53\%/47\% & 500 \\
\hline
CPSC Training & 6877 & 15.9 & 60.2 & 54\%/46\% & 500 \\
\hline
CPSC-Extra Train & 3453 & 15.9 & 63.7 & 53\%/46\% & 500 \\
\hline
Hidden CPSC & 2926 & 17.4 & 60.4 & 52\%/48\% & 500 \\
\hline
INCART & 72 & 1800.0 & 56.0 & 54\%/46\% & 257 \\
\hline
PTB & 516 & 110.8 & 56.3 & 73\%/27\% & 1000 \\
\hline
PTB-XL & 21837 & 10.0 & 59.8 & 52\%/48\% & 500 \\
\hline
G12EC (all data) & 20678 & 10.0 & 60.5 & 54\%/46\% & 500 \\
\hline
G12EC Training & 10344 & 10.0 & 60.5 &54\%/46\% & 500 \\
\hline
Hidden G12EC & 10344 & 10.0 & 60.5 & 54\%/46\% & 500 \\
\hline
Undisclosed & 10000 & 10.0 & 63.0 &53\%/47\% & 300 \\
\hline\hline
\end{tabular}%}
\end{center}
\caption{Relevant information for each data set}
\label{table:datasetsSummary}
\end{table*}

\subsubsection{Processing Pipeline fine-tuning}

\begin{figure*}[ht]
\centering
\includegraphics[width=\textwidth]{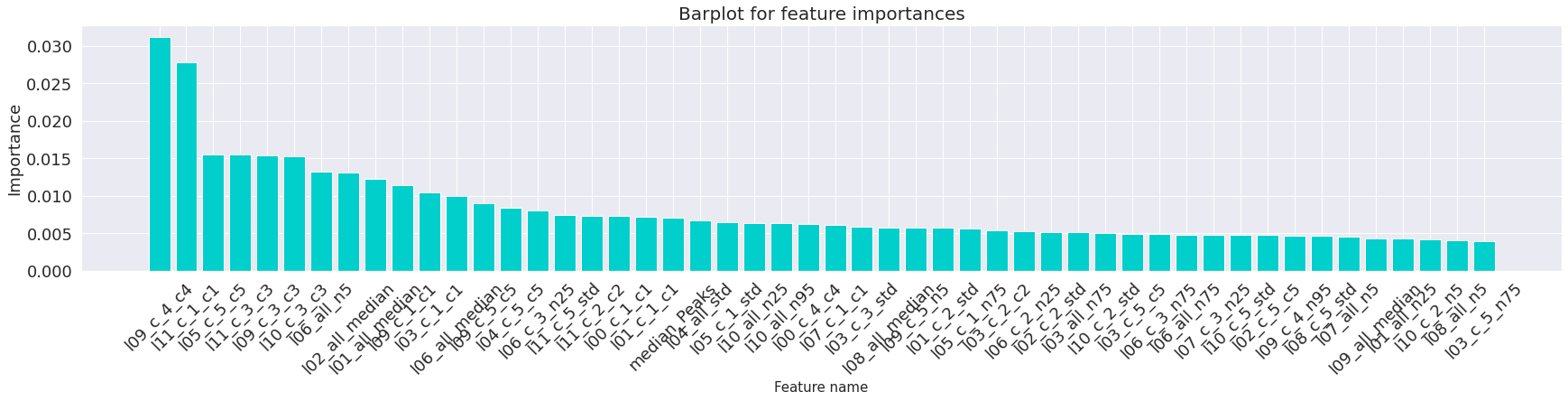}
\caption{Feature importance from XG-Boost algorithm (only the 50 best)}
\label{fig:feature_importance}
\end{figure*}

The bar plot in Fig.~\ref{fig:feature_importance} shows the most important features for predicting classes using the XG-Boost method. After obtaining importance scores for all 650 features, trial-and-error was conducted to remove unimportant features, ending with 120 features keeping the Accuracy and F1-Score. The best features included entropy for leads 9, 11, and 10, the lead six's 5th percentile, and the median for leads 1 and 2.

\begin{equation}
  feature_s = l(lead_i)\_c\_(coefficient_j)\_(operation_k)
  \label{eq:spectral_feature_detail}
\end{equation}

Each 636 spectral feature is based on \textit{lead} denoted as $l$, their \textit{coefficient} as $c$, and \textit{operation} applied on it. Also, $lead_i$ represents ECG lead number from 00-11, $coefficient_j$ represents a coefficient number from DWT (there are five coefficients in total 1-5), and $operation_k$ represents the operation name like average, standard deviation as std, variance as var, percentiles represents as n5(percentile 5), n25(percentile 25), n50(percentile 50), n75(percentile 75), n95(percentile 95) and entropy applied on coefficient got represented as $c1-c5$. In Fig.\ref{fig:feature_importance}, names of features are represented by the \ref{eq:spectral_feature_detail} mnemotechnic structure.

\begin{figure}[ht]
\centering
\begin{subfigure}{.5\textwidth}
  \centering
  \includegraphics[width=1\linewidth]{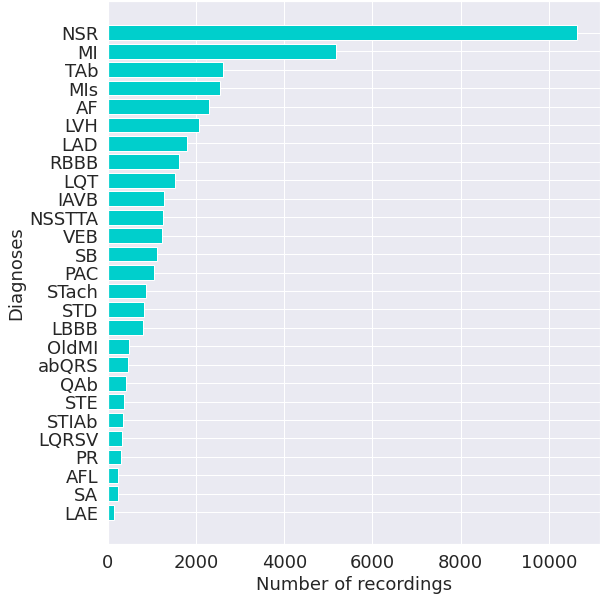}
  % \caption{}
  \label{fig:sub1}
\end{subfigure}%
\begin{subfigure}{.5\textwidth}
  \centering
  \includegraphics[width=1\linewidth]{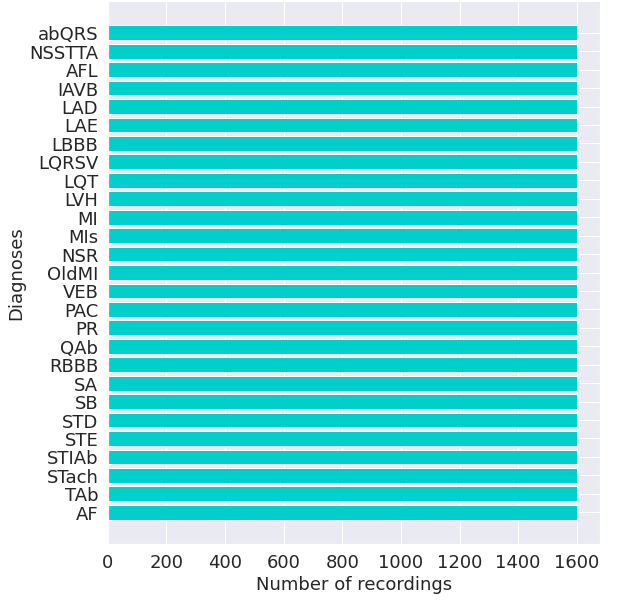}
  % \caption{}
  \label{fig:sub2}
\end{subfigure}
\caption{(Left) Selected diagnoses distribution. (Right) Diagnoses distribution for ROS and SMOTE datasets.}
\label{fig:recordingsPerDiagnose_append}
\end{figure}

The ROS and SMT approaches were applied to the dataset with a down-sampling to have as many recordings as the filtered dataset, with 43,200 recordings for each method. Thus, as depicted in Fig.~\ref{fig:recordingsPerDiagnose_append}, the distribution of the labels is much more similar among the arrhythmia categories.

\subsection{Scenarios}

\begin{figure*}[ht]
\centering
\includegraphics[width=\textwidth]{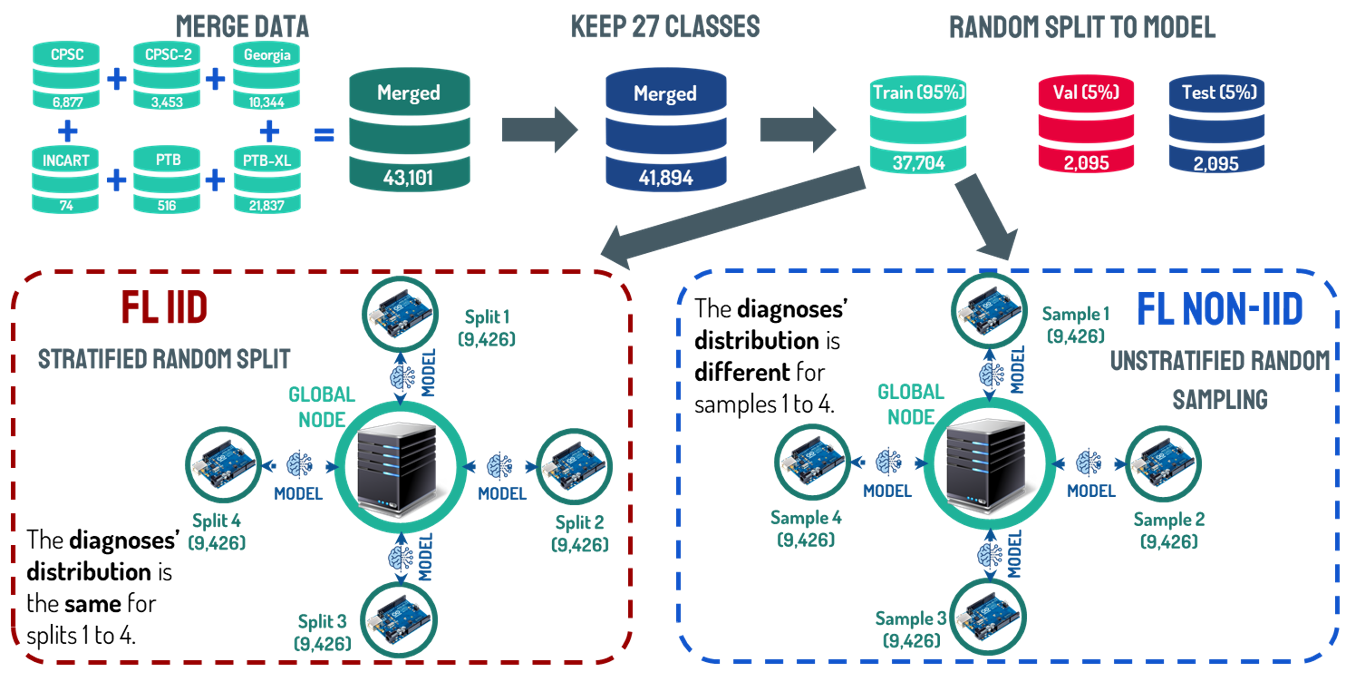}
\caption{IDD and Non-IID architecture}
\label{fig:fl_approach1}
\end{figure*}

We present an FL approach for classifying 12-channel ECG arrhythmias. Two FL scenarios, IID and Non-IID, were developed and tested (see Fig.~\ref{fig:fl_approach1}). The IID combined six databases into one dataset, resulting in 41,894 (a.k.a. filtered dataset) recordings after selecting the most representative diagnoses. The dataset was randomly split into train (90\%), validation (5\%), and test (5\%) sets. Stratified random splitting was used for the label distributions to be the same for each partition. Different partition sizes, such as 2, 4, 6, 8, and 10, are possible depending on the 12-leads ECG data, which will be explored in section \ref{sec:change_num_nodes}.

The Non-IID approach has the same two steps of the IID (data appending and splitting of Fig.~\ref{fig:fl_approach1}). The main difference is the generation of the local nodes. Thus, to get those clients, an Unstratified random sampling (with replacement) method~\cite{unstratified_rs} was employed. As a result, each local node has a varied distribution of diagnoses. Furthermore, some recordings may be missing entirely from the synthesized dataset.

\begin{figure*}[ht]
\centering
\includegraphics[width=\textwidth]{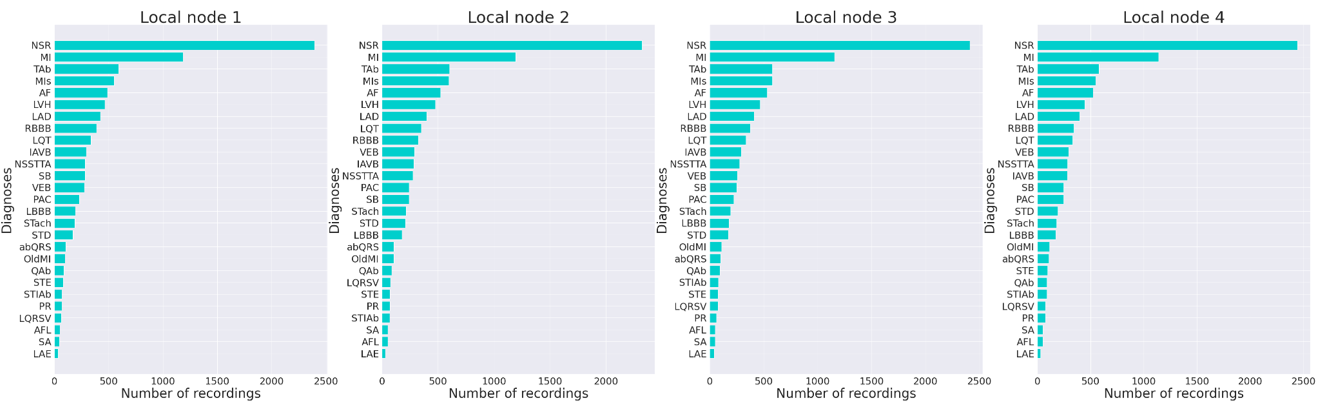}
\caption{Label distribution for the filtered dataset by each local node}
\label{fig:fl_label_distro_filtered}
\end{figure*}

\begin{figure*}[ht]
\centering
\includegraphics[width=\textwidth]{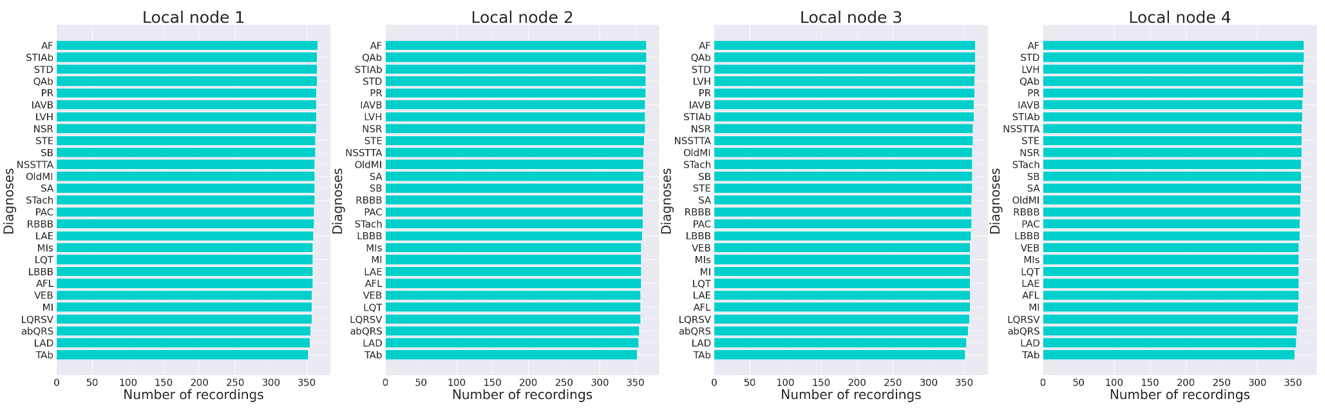}
\caption{Label distribution for the ROS and SMOTE datasets by each local node}
\label{fig:fl_label_distro_filtered_ROS_SMT}
\end{figure*}

As depicted in Fig.~\ref{fig:fl_label_distro_filtered}, the distribution among the four local nodes seems IID. The latter means that the diagnoses along the devices are equally distributed, containing 9,426 recordings per local node. The same occurs for the ROS and SMOTE datasets, as is shown in Fig.~\ref{fig:fl_label_distro_filtered_ROS_SMT}. 

\begin{figure*}[ht]
\centering
\includegraphics[width=\textwidth]{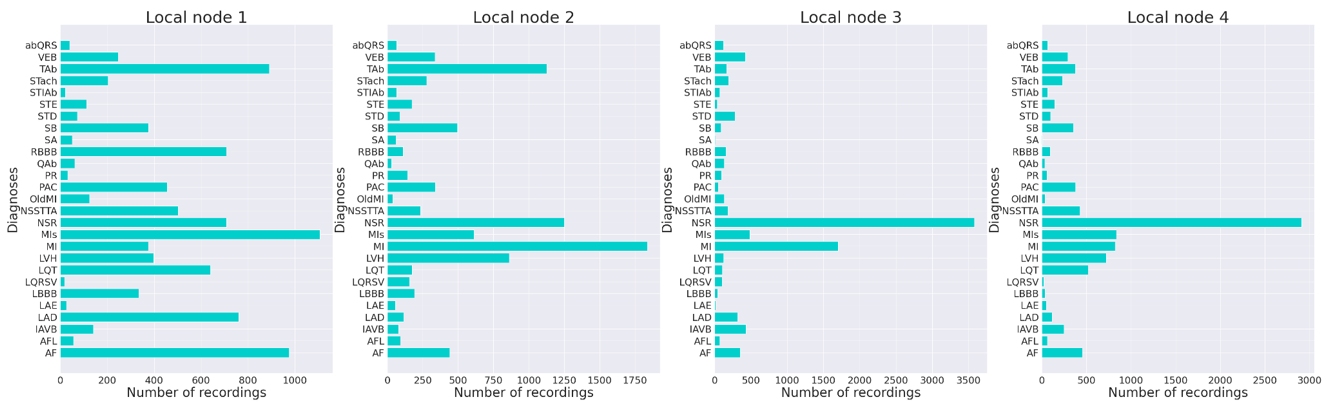}
\caption{Label distribution for the Non-IID method by local node}
\label{fig:fl_label_distro_filtered_noniid}
\end{figure*}

For the Non-IID approach, the method drew four random samples with repetition from the 41,894 recordings (local nodes) dataset. The given data in each local node (or client) is Non-IID approaches due to the unstratified random sampling, as shown in Fig.\ref{fig:fl_label_distro_filtered_noniid}. This scenario is far more plausible since most 12-channel ECGs shared across several devices are Non-IID.

\subsection{Performance metrics}

For measuring the goodness-of-fit for each method proposed, five metrics were employed. Those metrics are related to the concepts extracted from a confusion matrix \cite{metrics1}. The following are explanations about each term, using as an example a fictitious diagnosis A:

\begin{itemize}
    \item \textit{True negatives (TN):} The model predicted a patient wouldn't have diagnosis A, and he did not have it.
    \item \textit{True positives (TP):} These were examples when the model predicted yes (the recording has the diagnosis A), and the patient does have it.
    \item \textit{False positives (FP):} The model predicted that a patient would have the diagnosis A, but he does not. (also referred to as a "Type I error.")
    \item \textit{False negatives (FN):} The model predicted that a patient would not have diagnosed A, yet he does. (This is often referred to as a "Type II error.")
\end{itemize}

Then, the metrics to measure the classification power of the algorithms proposed are:

\textbf{Accuracy}: $Accuracy = \frac{TP+TN}{TP+FP+FN+TN}$: A metric measuring correct predictions out of all observations. Useful for balanced data where false positives and false negatives are similar.

% \begin{equation}
%     $Accuracy = frac{TP+TN}{TP+FP+FN+TN}$
% \end{equation}

\textbf{Precision}: $Precision = \frac{TP}{TP+FP}$: The ratio of accurately predicted positives to total expected positives. It reflects the accuracy of identifying patients with a specific diagnosis and relates to a low false-positive rate.
% \begin{equation}
%     Precision = \frac{TP}{TP+FP}
% \end{equation}

\textbf{Recall}: $Recall = \frac{TP}{TP+FN}$: The ratio of successfully predicted positives to all actual class observations. It measures the identification of patients with arrhythmia A.
% \begin{equation}
%     Recall = \frac{TP}{TP+FN}
% \end{equation}

\textbf{F1-Score}: $F1-Score = \frac{2*(Recall * Precision)}{(Recall + Precision)}$: A weighted average of Precision and Recall, considering false positives and false negatives. It is more valuable than accuracy, particularly when dealing with unequal class distribution or significant differences in the cost of false positives and negatives.
% \begin{equation}
%     F1 Score = \frac{2*(Recall * Precision)}{(Recall + Precision)}
% \end{equation}

It is relevant to highlight that the current work deals with a multi-class problem (more than two classes as labels). Due to it, for Precision, Recall, and F1-Score, the final metrics values were calculated as the weighted average of the metric measured in each pairwise class \cite{multiclass_metrics}.

In FL is relevant to measure the \textbf{Execution time} that a model takes to be trained \cite{fl24}. The latter helps compare the FL architecture to the centralized one, expecting the FL approach to be faster in each local node. This time accounts for the time elapsed since the model started to train until it finished and returned the results. During the experimentation of this work, the execution times were measured in \textbf{minutes}. The hardware specification used to train and evaluate the models is shown in Table~\ref{table:architecture_configxx}.

\begin{table}[ht]
\begin{center}
% \resizebox{7cm}{!}{
\begin{tabular}{ ||p{4cm}||p{5.2cm}|| }
 \hline
\textbf{Component} & \textbf{Specification} \\ [0.4ex] 
 \hline\hline
Disk size & 108 GB \\
\hline
Processors' model & Intel(R) Xeon(R) CPU @ 2.20GHz \\
\hline
Number of processors & 2 \\
\hline
Memory & 13.2 GB \\
\hline
Operating System & (Linux) Ubuntu 18.04.5 LTS \\
\hline
GPU & GeForce RTX 3070 8GB \\
\hline
Python version & 3.7.13 \\
\hline\hline
\end{tabular}%}
\end{center}
\caption{Hardware specification used to train models}
\label{table:architecture_configxx}
\end{table}

%%%%%%%%%%%%%%%%%%%%%%%%%%%%%%%%%%%%%%%%%
%%%%%%%%%%%%%%%%%%%%%%%%%%%%%%%%%%%%%%%%%
\section{Results}
\label{sec:results}

\subsection{Performance of Centralized}

A Centralized (traditional) Learning approach was employed to evaluate the dataset's performance without dividing it among hospitals and devices. Feature engineering, data balancing, and classification methods (DNN, LSTM) were applied to the complete dataset. The model of the second-best team (TEAM2)~\cite{competence_group2} in the Physionet competition was replicated for fair comparisons\footnote{TEAM2's code: \url{https://github.com/ZhaoZhibin/Physionet2020model}}. Imbalanced datasets for each arrhythmia type were addressed using ROS and SMT techniques. ROS was also applied to the TEAM2 strategy but did not yield better results. The SMT strategy was not tested within the TEAM2 approach. Table~\ref{table:scenarios_models} depicts the tested and best approaches we got regarding the classification metrics.

\begin{table}[ht]
\begin{center}
% \resizebox{7cm}{!}{
\begin{tabular}{ ||p{2.5cm}||p{7.0cm}||p{2.5cm}|| }
 \hline
\textbf{Characteristic} & \textbf{Scenarios} & \textbf{Best approach}\\ [0.4ex] 
 \hline\hline
 Data Split & \%Train-\%Validation-\%Test: \hspace{10 mm} Option 1: 60\%-20\%-20\% Option 2: 70\%-10\%-10\% Option 3: 80\%-10\%-10\% Option 4: 90\%-5\%-5\% & Option 4: 90\%-5\%-5\%\\
\hline
Features normalization & Option 1: MinMaxScaler \hspace{10 mm} Option 2: StandardScaler \hspace{10 mm} Option 3: RobustScaler & Option 3: RobustScaler \\
\hline
Sampling rate & Option 1: 257Hz \hspace{10 mm} Option 2: 500Hz & Option 1: 257Hz \\
\hline
Features employed & Option 1: Baseline features (morphological and demographic) \hspace{10 mm} Option 2: Baseline features (morphological and demographic) and Spectral features & Option 2: Baseline features and Spectral features \\
\hline\hline
\end{tabular}%}
\end{center}
\caption{Scenarios tested for centralized learning}
\label{table:scenarios_models}
\end{table}

\begin{figure}[ht]
\centering
\begin{subfigure}{.5\textwidth}
  \centering
  \includegraphics[width=1\linewidth]{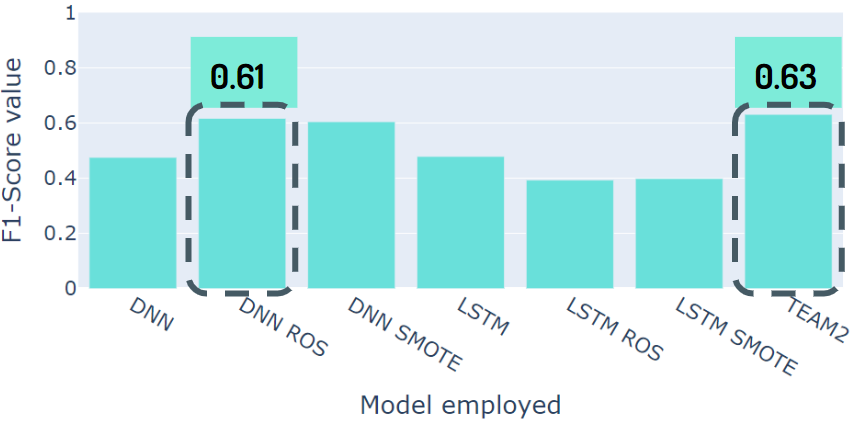}
  % \caption{}
  \label{fig:cl_time_f1score_methods_sub1}
\end{subfigure}%
\begin{subfigure}{.5\textwidth}
  \centering
  \includegraphics[width=1\linewidth]{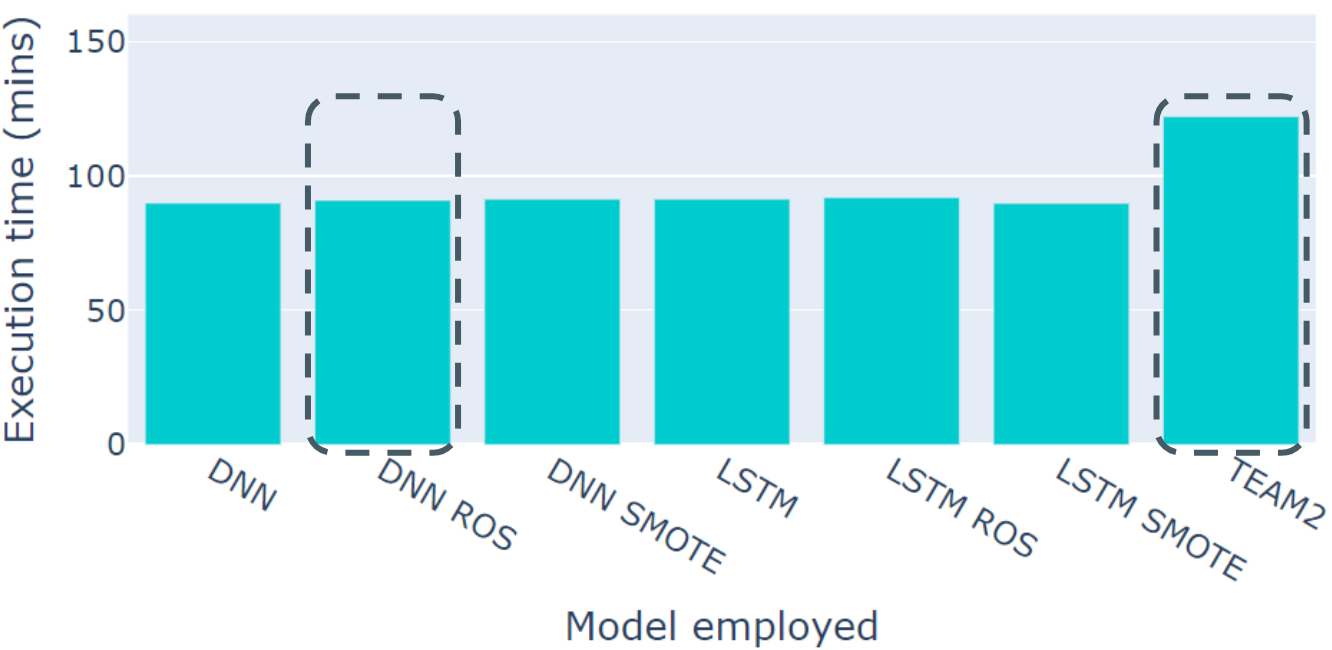}
  % \caption{}
  \label{fig:cl_time_f1score_methods_sub2}
\end{subfigure}
\caption{(Left) F1-Score for methods employed in CL on the test set. (Right) Execution times for the methods used in CL.}
\label{fig:cl_time_f1score_methods}
\end{figure}

Within the results in Fig.~\ref{fig:cl_time_f1score_methods}, the best model was TEAM2's approach, which got an F1-Score close to 0.63. Nevertheless, the DNN over the ROS data had similar behavior, with the mentioned metric at 0.61. Finally, LSTM does not perform well compared to the other models, with metrics between 0.39 and 0.47 for all the scenarios.

Finally, as shown in Fig.~\ref{fig:cl_time_f1score_methods}, TEAM2's approach took almost 122 minutes to run. On the other hand, DNN and LSTM took lower execution times (close to 89 minutes on average). Thus, TEAM2 is the slowest method, although it generates the best results. On the contrary, DNN is a fast method, and the performance is NOT quite different from TEAM2.

\subsection{Performance of IID Federated}

Then, the modeling part was executed in the IID FL paradigm using four clients. Therefore, DNN and LSTM methodologies classified the 12-leads ECG's arrhythmias. Hence, the aggregation technique employed to get the weights of the global model was FedAvg~\cite{fl19}.

\begin{figure}[ht]
\centering
\begin{subfigure}{.5\textwidth}
  \centering
  \includegraphics[width=0.9\linewidth]{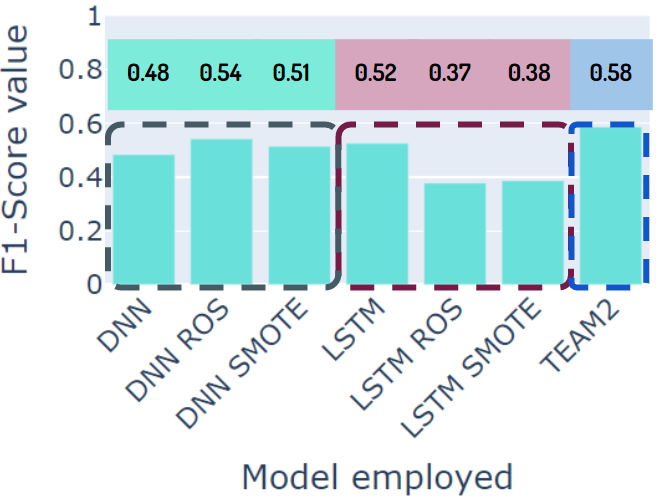}
  % \caption{}
  \label{fig:fl_time_iid_methods_sub1}
\end{subfigure}%
\begin{subfigure}{.5\textwidth}
  \centering
  \includegraphics[width=0.9\linewidth]{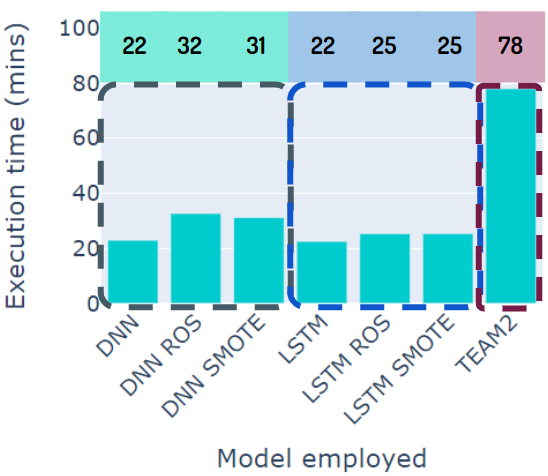}
  % \caption{}
  \label{fig:fl_time_iid_methods_sub2}
\end{subfigure}
\caption{(Left) F1-Score for methods employed in FL on the test set (IID). (Right) Execution times for the methods used in FL (IID).}
\label{fig:fl_time_iid_methods}
\end{figure}

% \begin{figure}[ht]
% \centering
% \includegraphics[width=0.6\textwidth]{paper-images/fl_iid_methods.png}
% \caption{F1-score for methods employed in FL on the test set}
% \label{fig:fl_iid_methods}
% \end{figure}

As depicted in Fig.~\ref{fig:fl_time_iid_methods}, the best results arise using the TEAM2 method with an F1-Score of 0.58. On the other hand, DNN got an F1-Score of 0.54, placing it as the second-best option. In addition, the best performance for LSTM was obtained with the original data, although it is worst than the TEAM2 and DNN ROS models. Then, applying oversampling for LSTM does not improve the result of the models. However, using ROS, the performance of the FL model increases. 

% \begin{figure}[ht]
% \centering
% \includegraphics[width=0.5\textwidth]{paper-images/times_fl_iid.png}
% \caption{Execution times for the methods used in FL IID approach}
% \label{fig:times_fl_iid}
% \end{figure}

Considering Fig.~\ref{fig:fl_time_iid_methods} results, the TEAM2 generates the slowest procedure with a time of 78 minutes. In addition, the second less time-consuming approach was the DNN ROS with 32 minutes, where using oversampling techniques makes the execution time increase~\cite{fl28}. Compared to the CL approach, the TEAM2 FL method has a faster execution with similar performance. Similarly, DNN ROS and SMOTE ran fast, performing comparably to their CL versions. 

\begin{figure*}[ht]
\centering
\includegraphics[width=\textwidth]{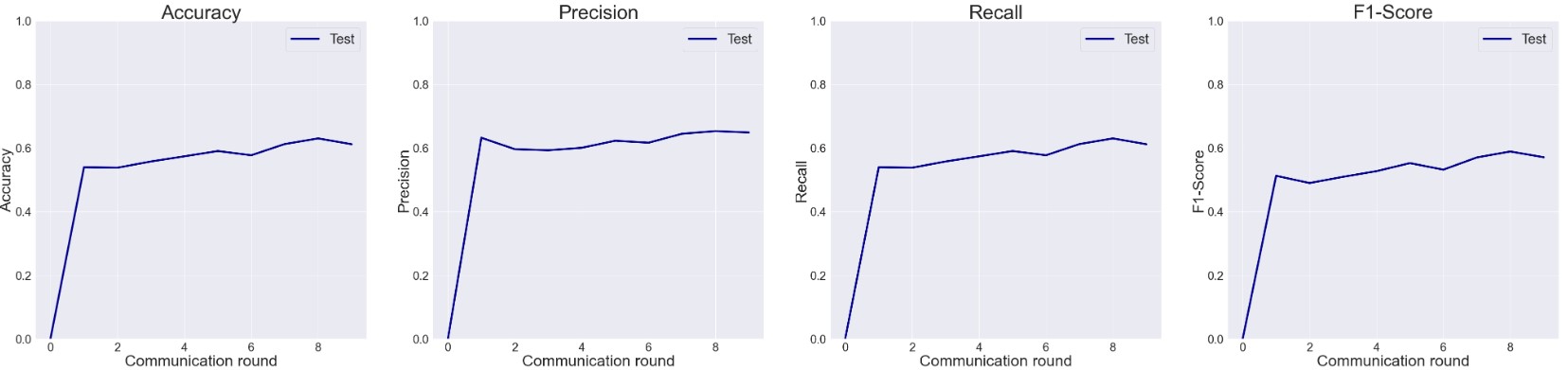}
\caption{Metrics along communication rounds for TEAM2 (the best model) on the test set}
\label{fig:comm_round_metrics_TEAM2}
\end{figure*}

Fig.~\ref{fig:comm_round_metrics_TEAM2} establishes the behavior of each metric for each communication round. As depicted, the performance stabilizes after the eighth comm round. Notice that all the metrics start low in the first comm round, but the measurements get steady after some updates. DNN and LSTM methods performed similarly, but for the sake of the extension, it is not included in this work.

\begin{table*}[ht]
\begin{center}
% \begin{tabular}{ ||p{1cm}||p{1cm}||p{1cm}||p{1cm}||p{1cm}||p{1cm}||}
% \resizebox{\width}{!}{
\begin{adjustbox}{width=\textwidth}
\begin{tabular}{|l|*{5}{c}|*{5}{c}|}

\cline{1-11} & \multicolumn{5}{c|}{CL} & \multicolumn{5}{c|}{FL IID}       \\  \cline{2-11}
 Method & \multicolumn{1}{c|}{Accuracy} & \multicolumn{1}{c|}{Precision}& \multicolumn{1}{c|}{Recall}& \multicolumn{1}{c|}{F1-Score} & \multicolumn{1}{c|}{Time} & \multicolumn{1}{c|}{Accuracy} & \multicolumn{1}{c|}{Precision}& \multicolumn{1}{c|}{Recall}& \multicolumn{1}{c|}{F1-Score} & \multicolumn{1}{c|}{Time} \\ 
 \hline \hline
\multicolumn{1}{|l|}{Competence Team \#2~\cite{competence_group2}} & \textbf{0.64} & 0.64 & \textbf{0.64} & \textbf{0.63} & 122 & \textbf{0.63} & \textbf{0.64} & \textbf{0.63} & \textbf{0.58} & 78 \\
\multicolumn{1}{|l|}{Inspirational DNN~\cite{inspiration_DNN}} & 0.50 & 0.46 & 0.50 & 0.47 & \textbf{88} & - & - & - & - & -  \\
\multicolumn{1}{|l|}{Inspirational LSTM~\cite{inspiration_DNN}} & 0.50 & 0.45 & 0.50 & 0.46 & 89 & - & - & - & - & -  \\
\hline\hline
% \multicolumn{1}{|l|}{XG-BOOST} & 0.52 & 0.49 & 0.52 & 0.49 & 9.4 & - & - & - & - & -  \\
% \multicolumn{1}{|l|}{XG-BOOST ROS} & 0.45 & 0.58 & 0.45 & 0.47 & 15.7 & - & - & - & - & - \\ 
% \multicolumn{1}{|l|}{XG-BOOST SMOTE} & 0.48 & 0.55 & 0.48 & 0.49 & 15.4 & - & - & - & - & -  \\
% \multicolumn{1}{|l|}{CATBOOST} & 0.55 & 0.54 & 0.55 & 0.53 & 34.5 & - & - & - & - & - \\
% \multicolumn{1}{|l|}{CATBOOST ROS} & \textbf{0.67} & \textbf{0.72} & 0.67 & \textbf{0.67} & 36.4 & - & - & - & - & -  \\
% \multicolumn{1}{|l|}{CATBOOST SMOTE} & 0.64 & 0.66 & 0.64 & 0.64 & 35.5 & - & - & - & - & -  \\
% \hline\hline
\multicolumn{1}{|l|}{DNN} & 0.50 & 0.47 & 0.50 & 0.47 & 89 & 0.46 & 0.53 & 0.45 & 0.49 & \textbf{22}  \\
\multicolumn{1}{|l|}{DNN ROS} & 0.61 & \textbf{0.66} & 0.61 & 0.61 & 90 & 0.55 & 0.59 & 0.54 & 0.54 & 32  \\
\multicolumn{1}{|l|}{DNN SMOTE} & 0.60 & 0.64 & 0.60 & 0.60 & 91 & 0.52 & 0.58 & 0.52 & 0.51 & 31  \\
\hline\hline
\multicolumn{1}{|l|}{LSTM} & 0.51 & 0.47 & 0.51 & 0.39 & 91 & 0.48 & 0.58 & 0.48 & 0.52 & \textbf{22}  \\ 
\multicolumn{1}{|l|}{LSTM ROS} & 0.38 & 0.51 & 0.38 & 0.39 & 91 & 0.39 & 0.48 & 0.38 & 0.38 & 25  \\
\multicolumn{1}{|l|}{LSTM SMOTE} & 0.39 & 0.49 & 0.39 & 0.39 & 89 & 0.39 & 0.47 & 0.39 & 0.38 & 25  \\
\hline
\end{tabular}
% }
\end{adjustbox}
\end{center}
\caption{Metrics for CL and IID Federated Learning (FL -IID) in the test data set. Execution time for preprocessing and training in minutes.}
\label{table:metrics_CL_FLIID}
\end{table*}

Table~\ref{table:metrics_CL_FLIID} depicts the metrics obtained over the test dataset and the execution time of the training phase and all the state-of-the-art competitors. The CL TEAM2 model outperforms all the proposals by at least two percentage points, but it is the slowest method. Concerning the FL architecture, the TEAM2 FL is close enough to the centralized TEAM2 model, showing that the FL applied over an IID set behaves well, although the execution time is better for DNN ROS.

\subsection{Performance of Non-IID Federated}

The modeling phase was done again with the four Non-IID datasets extracted. The 12-channel ECG arrhythmias were then classified using the TEAM2, DNN, and LSTM techniques following an FL paradigm.

\begin{figure}[ht]
\centering
\begin{subfigure}{.5\textwidth}
  \centering
  \includegraphics[width=1\linewidth]{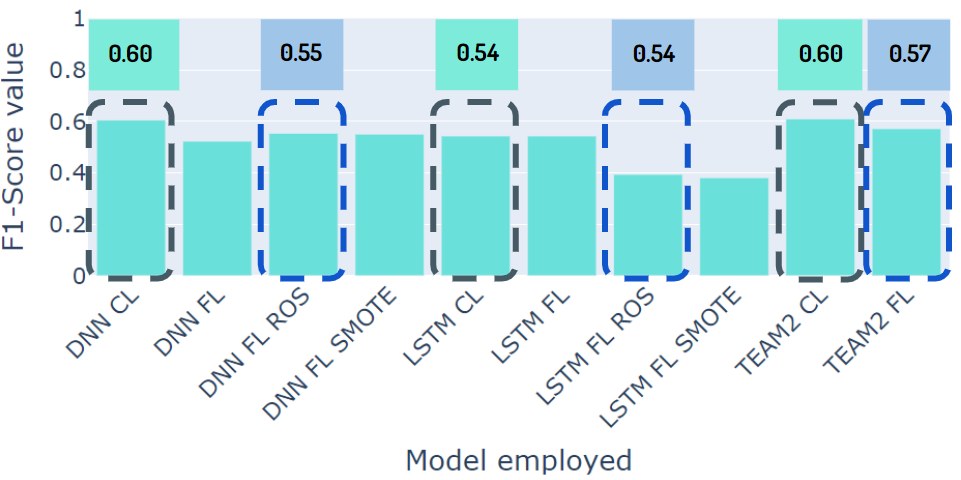}
  % \caption{}
  \label{fig:fl_time_noniid_methods_sub1}
\end{subfigure}%
\begin{subfigure}{.5\textwidth}
  \centering
  \includegraphics[width=1\linewidth]{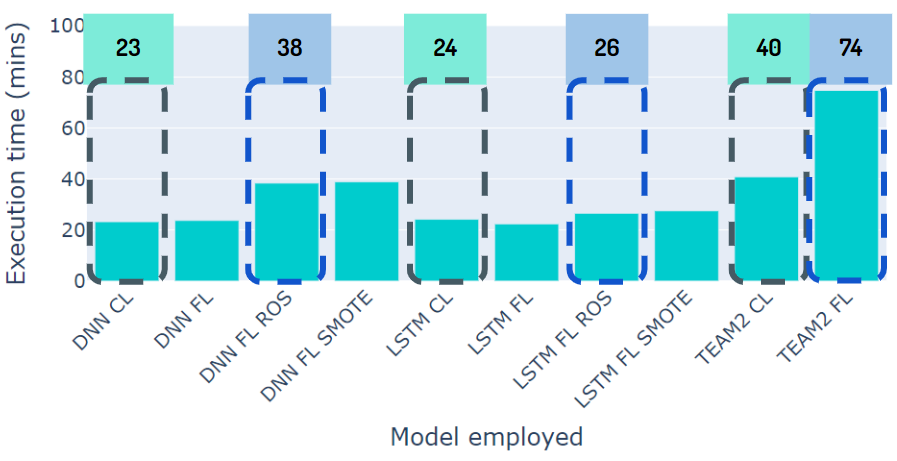}
  % \caption{}
  \label{fig:fl_time_noniid_methods_sub2}
\end{subfigure}
\caption{(Left) F1-Score for methods employed in FL Non-IID on the test set. (Right) Execution times for the methods used in FL Non-IID.}
\label{fig:fl_time_noniid_methods}
\end{figure}

% \begin{figure}[ht]
% \centering
% \includegraphics[width=0.8\columnwidth]{paper-images/fl_noniid_methods.png}
% \caption{F1-score for methods employed in FL Non-IID on the test set}
% \label{fig:fl_noniid_methods}
% \end{figure}

We employed a CL strategy to compare the Non-IID FL and CL implementations. In terms of classification performance, the best results were for TEAM2 with the original datasets, as shown in Fig.~\ref{fig:fl_time_noniid_methods}, while the DNN ROS and DNN LSTM techniques also performed well, with an F1-Score of around 0.58 on the test set. On the other hand, the highest performance for LSTM was with the original data, yet it was worse than the DNN ROS model. 
% Then, using oversampling approaches does not improve the results of the LSTM models. 

% \begin{figure}[ht]
% \centering
% \includegraphics[width=0.8\columnwidth]{paper-images/times_fl_noniid.png}
% \caption{Execution times for the methods used in FL Non-IID approach}
% \label{fig:times_fl_noniid}
% \end{figure}

As shown in Fig.~\ref{fig:fl_time_noniid_methods}, the TEAM2 method was the slowest, clocking in at 74 minutes. Compared to the CL method, the TEAM2 methodology with FL is faster and produces equivalent results. The same behavior applies to DNN ROS and SMOTE; however, while LSTM performs worse than TEAM2 and DNN, it is significantly quicker in producing the results.

\subsection{Performance changing local nodes} \label{sec:change_num_nodes}

We implemented a model's performance evaluation by changing the number of local nodes diverse to 4 (i.e., 2, 4, 6, 8, 10). Per each client, we trained a CL to determine how well the FL training fits regarding that CL for IID and Non-IID scenarios. Still, for the sake of the extension of this document, only the Non-IID result is reported because the performances of both methods were quite similar.

\begin{figure}[ht]
\centering
\begin{subfigure}{.5\textwidth}
  \centering
  \includegraphics[width=1\linewidth]{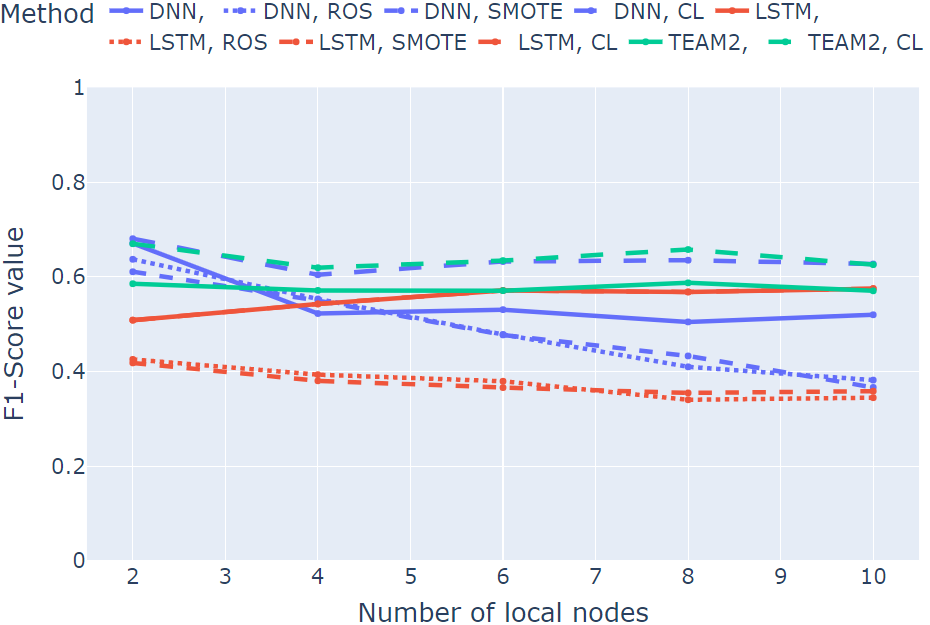}
  % \caption{}
  \label{fig:change_local_nodes_time_metrics_fedavg_sub1}
\end{subfigure}%
\begin{subfigure}{.5\textwidth}
  \centering
  \includegraphics[width=1\linewidth]{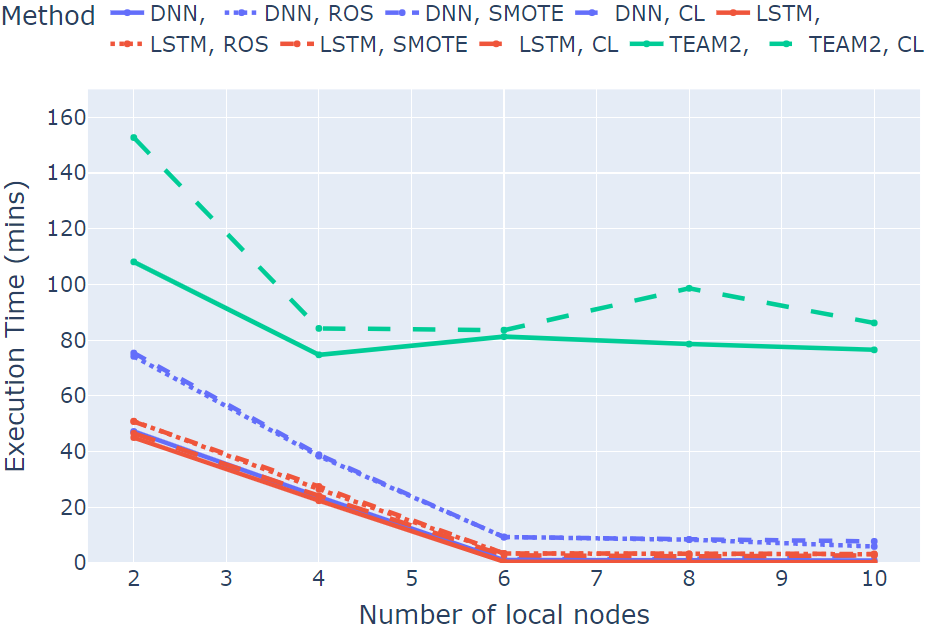}
  % \caption{}
  \label{fig:change_local_nodes_time_metrics_fedavg_sub2}
\end{subfigure}
\caption{(Left) F1-Score changing number of local nodes with FedAvg in test set. (Right) Running time changing number of local nodes with FedAvg.}
\label{fig:change_local_nodes_time_metrics_fedavg}
\end{figure}

% \begin{figure}[H]
% \centering
% \includegraphics[width=0.8\columnwidth]{paper-images/change_local_nodes_metrics_fedavg.png}
% \caption{F1-score changing number of local nodes with FedAvg in test set}
% \label{fig:change_local_nodes_metrics_fedavg}
% \end{figure}

Fig.~\ref{fig:change_local_nodes_time_metrics_fedavg} represents the F1-Score for all the methods while changing the number of local nodes. TEAM2 CL, DNN CL, and LSTM CL are reference models trained with all the data appended to a single dataset, where the best solution when considering two local nodes is DNN ROS. On the other hand, with four or more clients best solution is TEAM2, keeping steady behavior when increasing the number of clients. Nevertheless, when considering six or more nodes, LSTM on the original data is a good option, similar to the TEAM2 metric. 

% \begin{figure}[H]
% \centering
% \includegraphics[width=0.8\columnwidth]{paper-images/change_local_nodes_time_fedavg.png}
% \caption{Running time changing number of local nodes with FedAvg}
% \label{fig:change_local_nodes_time_fedavg}
% \end{figure}

Typically, the higher the local nodes' quantity, the faster the algorithms run, as depicted in Fig.~\ref{fig:change_local_nodes_time_metrics_fedavg}. Moreover, all the techniques drastically decreased the running time when increasing the number of clients, but the performance also decreased.

%%%%%%%%%%%%%%%%%%%%%%%%%%%%%%%%%%%%%%%%%
%%%%%%%%%%%%%%%%%%%%%%%%%%%%%%%%%%%%%%%%%
\section{Discussion}
\label{sec:disc}

% Which model performs better? Why?

Analyzing the results obtained from the experiments in the CL setting, the highest classification performance was reached by TEAM2, which combined Convolutional Recurrent Neural Networks (CRNN) and cross-validation. Nevertheless, our proposed DNN on the augmented ROS dataset had a competitive classification performance. In addition, the execution time for DNN was more than twice faster than TEAM2. \textbf{Why can a DNN model perform similarly to a CRNN while reducing the execution time?} This is possible since we employed a solid pre-processing pipeline that extracted and selected powerful spectral and morphological features to classify arrhythmia accurately. Moreover, the architecture of DNN was considerably less complex (fewer hidden layers, fewer hidden neurons, not using cross-validation) than TEAM2's approach.

Concerning the FL IID experiments results, TEAM2 and DNN using the ROS balanced data have similar F1-Score to their CL versions while having a decrease of around 36\% and 64\% in the execution time, respectively. \textbf{Why does employing FL cause such execution time reduction?} That is an expected behavior supporting scientific evidence~\cite{abdulrahman2020survey} since running models in an FL fashion can be done parallelly in each client, removing the burden of transmitting large amounts of data, and causing the execution time to drop substantially.

Regarding the FL non-IID experiment results, TEAM2 and DNN using ROS again reached a classification score similar to their CL approaches, causing an increase of 85\% and 65\% in the execution time, respectively. \textbf{How come that FL IID causes a decrease in the execution time, but FL non-IID increases it?} The answer to this question involves the aggregation method employed (FedAvg) and the diversity of the data among the clients. Given that the data is non-IID by construction, it causes the time taken for the models to converge to be more significant since, in each communication round, the weights to aggregate are much more different from one another, which is something that does not occur in the IID (and not quite real) setting.

In the context of FL Non-IID, using less than four clients, the most accurate models were DNN and TEAM2. Even so, when considering six or more clients, LSTM exhibited a comparable performance while drastically decreasing the execution time. \textbf{Why does the number of clients involved in the FL experiment cause LSTM to outperform DNN and become more similar to TEAM2's approach?} The FL approach under DNN may overfit when training with fewer data in each client since the model becomes too specialized for a smaller dataset. At the same time, the LSTM algorithm has been studied as a more robust solution to overfitting (mitigating the vanishing gradient problem) \cite{ookura2020efficient}. In this case, the model is not as complex as the DNN, having a better performance while increasing the number of clients (decreasing the number of samples per client). CRNN may also overfit the given data, but that effect is mitigated by using cross-validation in TEAM2's solution, causing a comparable performance between TEAM2 and LSTM.

%%%%%%%%%%%%%%%%%%%%%%%%%%%%%%%%%%%%%%%%%
%%%%%%%%%%%%%%%%%%%%%%%%%%%%%%%%%%%%%%%%%
\section{Conclusions and future work}
\label{sec:conclusions}
Our team implemented a 12-lead arrhythmia classification leveraged on an FL paradigm over multiple ML and AI algorithms to evaluate their performance compared to a centralized (traditional) approach. Although the solution from TEAM2 provided good results, our proposal, using DNN over augmented datasets (ROS), produced a comparable performance with the benefit of a drastic reduction in the execution time. In addition, considering a Non-IID scenario over a few clients, TEAM2 and DNN ROS attained equivalent behaviors to their respective CL and FL approaches. At the same time, our DNN ROS solution acutely reduced the execution time. Alternatively, using our proposed LSTM for larger clients demonstrated steady and analogous behavior to the obtained in a CL with a drastic reduction of the execution time compared to the state-of-the-art.

% What else do we want to test?

In future work, the FL architecture could use a different data partition. For example, each original database can be used as a client with a Non-IID distribution of arrhythmias. Moreover, alternative aggregation methods may deal with the Non-IID property, such as Fedprox, SCAFFOLD, or FedNova~\cite{fl19}. Another venue for future research is using Catboost and XG-Boost in an FL architecture to check their performance for the 12-leads ECG arrhythmia classification.

\section*{Acknowledgements}
This work was partially supported by project SERICS (PE00000014) under the MUR National Recovery and Resilience Plan funded by the European Union - NextGenerationEU and PNRR351 TECHNOPOLE - NEXT GEN EU Roma Technopole - Digital Transition, FP2 - Energy transition and digital transition in urban regeneration and construction.

% This work was partially supported by project SIAE - Gestione dei diritti d’autore su reti 5G con Blockchain and by the program PNRR351 TECHNOPOLE - NEXT GEN EU Roma Technopole - Digital Transition, FP2 - Energy transition and digital transition in urban regeneration and construction.
\bibliographystyle{splncs04}
\bibliography{mybibliography}




% LNCS Homepage, \url{http://www.springer.com/lncs}. Last accessed 4
% \end{thebibliography}
% \appendix
% \label{sec:append}
% \input{sec-append}

\end{document}